\documentclass[journal]{IEEEtran}
\usepackage[utf8]{inputenc}
\usepackage{graphics} 
\usepackage{epsfig} 
\usepackage{mathptmx} 
\usepackage{times} 
\usepackage{wrapfig}

\usepackage{xcolor}

\newcommand{\argmax}{\operatornamewithlimits{argmax}}

\newcommand{\E}{\mathbb{E}}

\newcommand{\sinc}{sinc}

\usepackage{relsize}
\usepackage{enumerate}   

\usepackage{amsmath,amsthm,amssymb,amsfonts,dsfont} 
\usepackage{mathtools,thmtools}
\usepackage{algorithm,algorithmicx,listings} 
\usepackage[noend]{algpseudocode} 
\usepackage{graphicx}
\usepackage{subfig}
\usepackage[font={small}]{caption}
\usepackage{hyperref}
\usepackage{comment}

\title{Learning to Track Dynamic Targets in Partially Known Environments}
\author{Heejin Jeong$^{1}$, Hamed Hassani$^{1}$, Manfred Morari$^{1}$, Daniel D. Lee$^{2}$ and George J. Pappas$^{1}$
\thanks{$^{1}$These authors are with the Department of Electrical and Systems Engineering, University of Pennsylvania, Philadelphia, PA 19104, USA (e-mail:
        \{heejinj, hassani, morari, pappasg\}@seas.upenn.edu)}%
\thanks{$^{2}$Daniel D. Lee is with the Department of Electrical and Computer Engineering, Cornell University, Ithaca, NY 14850, USA (e-mail: dd146@cornell.edu)}%
         \thanks{This research is partially supported by ARL CRA DCIST W911NF-17-2-0181.}
}

\begin{document}

\maketitle
\begin{abstract}
We solve active target tracking, one of the essential tasks in autonomous systems, using a deep reinforcement learning (RL) approach. In this problem, an autonomous agent is tasked with acquiring information about targets of interests using its on-board sensors. The classical challenges in this problem are system model dependence and the difficulty of computing information-theoretic cost functions for a long planning horizon. RL provides solutions for these challenges as the length of its effective planning horizon does not affect the computational complexity, and it drops the strong dependency of an algorithm on system models. In particular, we introduce Active Tracking Target Network (ATTN), a unified RL policy that is capable of solving major sub-tasks of active target tracking -- in-sight tracking, navigation, and exploration. The policy shows robust behavior for tracking agile and anomalous targets with a partially known target model. Additionally, the same policy is able to navigate in obstacle environments to reach distant targets as well as explore the environment when targets are positioned in unexpected locations.
\end{abstract}
\section{Introduction} \label{sec:intro}
Active target tracking is an information gathering task where a mobile agent makes a sequence of control decisions to gather information about targets of interest using its on-board sensors. For instance, a camera angle can be controlled to find an exit in a room, or an autonomous vehicle can move to view and identify an occluded object. Its various applications include surveillance \cite{rybski00, hilal13}, search-and-rescue task \cite{kumar04}, active perception \cite{soatto11, zhang17, luo17, jayaraman18}, and environmental monitoring \cite{dunbabin12, choi09}. 

Problems in active tracking for dynamic targets tackle challenges associated with target motions such as estimation of target states, non-myopic planning, cost function, and stochasticity of the targets, rather than placing the weight on processing and inferencing sophisticated raw sensory data. Previous studies applied various techniques from Bayesian estimation, information theory, and optimal control to solve a spectrum of problems. 
However, many approaches often face a significant challenge in computational complexity due to their dependency on a planning horizon \cite{chung06, kreucher05, huber09}. Especially, it is often taxing to compute an information-theoretic cost function such as mutual information and variance reduction over a long planning horizon. Myopic policies have been proposed to lessen the computational burden, but often resulted in inefficient planning trajectories or failed to achieve the optimal performance. For example, short paths that are searched or sampled by a myopic planning approach fail to collect new information about a distant target, and thus such approaches struggle to find an optimal path until an agent gets close enough to the target. To improve the limitation, approximate non-myopic algorithms are presented and they reduce the complexity of obtaining a sub-optimal policy while providing strong performance guarantees \cite{atanasov14, schlotfeldt18}. An iterative sampling-based algorithm was also proposed, increasing efficiency under a given budget constraints \cite{hollinger2014sampling}. The complexity of their methods, however, are still not free of the planning horizon.

Another major limitation of previous works is a strong dependency of their methods on a specific problem formulation and system models. Due to such dependencies, they have focused on settings where the prior knowledge of targets (e.g. initial states and target dynamic models) are sufficiently known and/or the targets are initially located near the sensing range \cite{chung06, atanasov14, schlotfeldt18}. However, such conditions considerably simplify the problem as the main task becomes tightly following a target whose state is relatively accurately estimated. In real-world examples, we often have limited information about targets. For instance, an initial belief about a target can be inaccurate, leading to issues when the agent searches for the target near its incorrect belief location. Such scenarios require the agent to explore the environment until it discovers the targets. 
These limitations can be also problematic when targets move quickly or when their states considerably differ from predicted states by the agent. The latter is caused by noisy target dynamics or insufficient target knowledge available to the agent. 

In this paper, we address these main challenges and consider three major sub-tasks of active target tracking -- \textit{navigation}, \textit{discovery}, and \textit{in-sight tracking} (tracking within the range of a sensor). We propose a method of learning a unified reinforcement learning (RL) policy that is capable of all three tasks. We formulate the active target tracking problem as a Markov decision process (MDP) with an information-theoretic objective function where the value function is a discounted sum of future mutual information between target states and measurement history. An agent explicitly maintains a belief distribution of a target, and the belief state and its uncertainty are included in an RL state. One of the major advantages of using RL is its non-myopic behavior by its nature. A learned policy is executable online and its computational complexity is independent of the planning horizon. Another benefit is that it can drop the dependency of an algorithm on prior conditions and system models -- agent dynamics, target dynamics, and observation model. Therefore, it is able to learn a \textit{robust} behavior for a scenario where the agent's knowledge of the target model considerably differs from a true target model. We additionally incorporate map information and visit-frequency, which reflects how recently a certain area has been scanned by the sensor, into an RL state as images. This enables an RL policy to \textit{learn to navigate} and \textit{learn to explore}. Our model uses egocentric information with respect to the agent frame to drop the dependency of a learned policy on its training environment. We demonstrate the proposed method in a target tracking environment with various scenarios including agile targets, imperfect prior knowledge on a target model, and inaccurate initial beliefs. The results show that the RL policy significantly outperforms an existing search-based planning method.

A preliminary version of this paper appeared at the IEEE/RSJ International Conference on Intelligent Robots and Systems (IROS) 2019 \cite{jeong19_iros}. The paper proposed a general framework for applying reinforcement learning to active information acquisition and demonstrated the method in an active target tracking problem using model-free deep reinforcement learning algorithms. This article presents a new method of learning a unified policy for active target tracking that improves the previous approach by utilizing map information and visit frequency as inputs to an RL policy. We also demonstrate the capability of the method in partially known environments. This extends the problem domain to a more highly complex and advanced level.
\begin{figure*}[t!]
    \centering
    \includegraphics[width=0.8\textwidth]{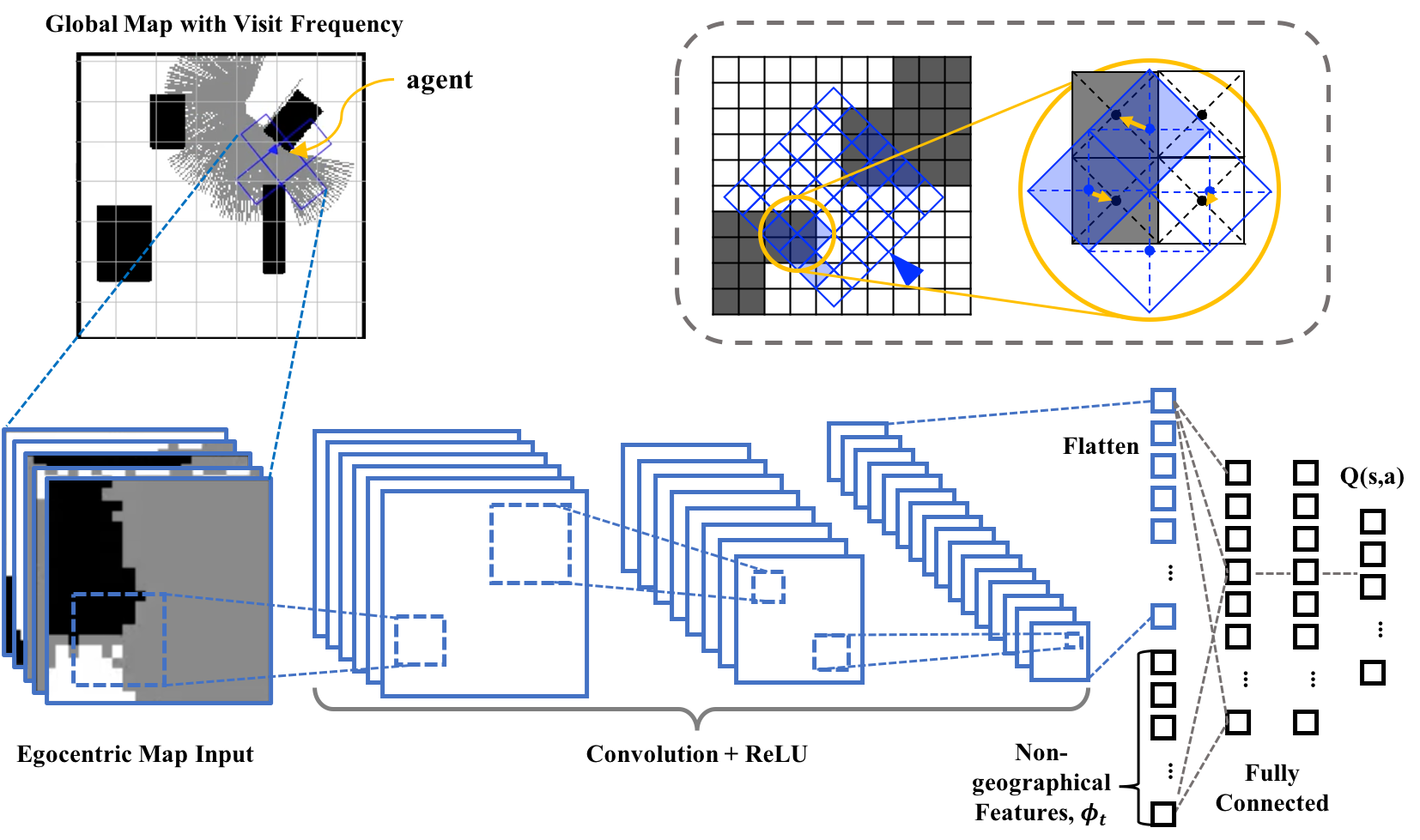}
    \caption{Illustration of the active target tracking network (ATTN) architecture. Egocentric maps with visit frequency of surroundings are fed to the convolutional neural network, and its output is concatenated by non-geographical features and finally estimates Q-values.}
    \label{fig:architecture}
\end{figure*}
\section{Related Work} \label{sec:related_work}
The field of active target tracking has been explored by various domains such as state estimation, sensor management, active perception, planning, and machine learning. Previous works that are closely related to this study are search-based target tracking algorithms for dynamic targets, Reduced Value Iteration (RVI) and Anytime Reduced Value Iteration (ARVI) \cite{atanasov14,schlotfeldt18}. Both algorithms formulate the active information acquisition problem as a deterministic optimal control problem and compute an open-loop policy that maximizes a mutual information objective. With known system models of an agent, targets, and sensors (observation), they build a search tree of the possible trajectories of the agent and apply a pruning method to reduce the size of the search tree and to ensure finite execution time while guaranteeing suboptimality. ARVI improves RVI by eliminating the need to tune pruning parameters in RVI while reducing the computation time by re-using computations from prior iterations. Although the algorithm shows promising results in simulation and is demonstrated in real robot experiments, it requires adequate knowledge of a true target model. Since the search tree is built with a predicted target trajectory by a Kalman filter, inaccurate knowledge of the target model can lead to a path where the target does not exist. In many cases, however, target models may not be known resulting in practical limitations.

Learning-based methods have not been extensively studied in the problems of dynamic targets, but numerous studies have integrated machine learning methods into active tracking tasks of static objects. They mainly focus on complications of target objects such as arbitrarily deforming objects, self-occluded objects, and semantic understanding \cite{soatto11, motchallenge, giusti15}. The recent advancements in deep learning in computer vision and perception enable solutions to track more sophisticated objects or to study complex scenarios. Following the trend, some of the recent literature in active perception apply deep RL methods to directly learn a control policy from raw sensory inputs \cite{whitehead90, zhang17, luo17, jayaraman18}. Active target tracking can be seen as a sequential decision making problem and formulated as an MDP. RL has been widely used to learn an optimal policy for such sequential decision making problems when knowledge on models is not available or too complicated to solve using exhaustive methods such as dynamic programming.  Zhang et. al. proposed a RL method for visual tracking in videos that uses REINFORCE \cite{williams1992simple} and a recurrent convolutional neural network. Jayaraman et. al. also proposed a deep RL solution which learns an exploratory behavior for active completion of panoramic natural scenes and 3D object shapes.

Exploratory behavior for information gathering has long been studied in robotics. Various approaches have been presented including heuristic methods \cite{yamauchi1997frontier,gonzalez2002navigation}, formal decision theory \cite{amigoni2010information}, and planning \cite{bourgault, kollar2008trajectory, lauri2016planning}. Many of the studies aim to increase the accuracy in building maps in unknown environments. Myopic approach is studied which maximizes the expected Shannon information gain of an occupancy map as well as minimizes the uncertainty of the vehicle pose and map feature uncertainty \cite{bourgault}. It is also solved as a sequential decision making problem, formulated as a partially observable Markov decision process (POMDP). Kollar and Roy \cite{kollar2008trajectory} applied policy search dynamic programming \cite{bagnell04} to find an optimal trajectory for improving accuracy of a map by minimizing the number of sensor measurements for map features. An approach proposed in \cite{lauri2016planning} solves a POMDP with sequential Monte Carlo optimization. They use a sampling-based approximate mutual information for its reward function. More recently, learning-based approaches have been presented to solve exploration tasks. Chen et. al. used imitation learning to pre-train an exploration policy, and then applied a policy gradient algorithm to further optimize the policy. Similar to our approach, they use egocentric maps for the network inputs. Their reward function includes a coverage measure as well as a collision penalty with fine-tuned parameters. Our goal for exploration is to find dynamic objects, and differs from these previous works which aim to construct accurate information about static surroundings. 

Navigation tasks of mobile robots have also been paid increasing attention from the learning community \cite{gupta17, kahn17, giusti15}. Gupta et. al. proposed an architecture that builds a belief map of a world and trains a mapper and a planner to reach a goal \cite{gupta17}. However, one of the challenges for applying such navigation approaches to target tracking scenarios is that a goal position continuously changes as targets move. Re-planning its path for a new goal every step is computationally expensive. Another challenge is that a goal position is not always identical to a belief position. For instance, in a situation with two targets, the optimal goal position for the agent is located somewhere in between the targets and is also dependent on their uncertainties.

\section{Approach} \label{sec:approach}
\subsection{Problem Formulation}
We are interested in an active target tracking problem where both an agent and a target are mobile following discrete-time dynamic models. We consider one robot and $N$ targets, and denote their states at time $t$ as $x_t$ and $y_{t} = [y_{1,t}^T, \cdots, y_{N,t}^T]^T$ where $y_{i,t}$ is an individual target state for $i=1, \cdots, N$. The robot state can be defined in a different form depending on its degree of freedom or its type. For instance, $x_t \in \mathbb{R}^2 \times SO(2)$ for a ground robot and $x_t\in \mathbb{R}^3 \times SO(3)$ for a quadrotor where $SO(n)$ is the special orthogonal group in dimension $n$. The goal of the robot is to maximize the cumulative mutual information between the belief states at $t$ and a measurement history $z_{1:t}$ for a time horizon $T$. More formally, the objective is to find a sequence of control inputs to the robot, $u_{1:T}$, which satisfies, 
\begin{equation}
    \text{maximize}_{u_{0:T-1}} \sum_{t=1,\cdots, T} \mathrm{I}(y_{t}; z_{1:t}|x_{1:t})
    \label{eq:objective}
\end{equation}
	\begin{align*}
		    s.t.\quad & x_{t+1} = f(x_t, u_t) & t=0,\cdots, T-1\\
		        & y_{i,t+1} = g(y_{i,t}) & t=0,\cdots, T-1 \\
		        & z_{i,t} = h(x_t, y_{i,t}) & t=1, \cdots, T
		\end{align*}
where $f(\cdot)$ and $g(\cdot)$ are dynamic models of the robot and the targets, respectively, and $h(\cdot)$ is an observation model.    
Since the robot does not have access to the ground truth of the target states, the robot infers the target states from its internal belief distributions. We denote the belief distribution for the $i$-th target as $B(y_{i,t}) = p(y_{i,t}|z_{1:t}, x_{1:t})$ and its predicted distribution for the subsequent step as $\bar{B}(y_{i,t+1})=p(y_{i,t+1}|z_{1:t},x_{1:t})$. 

Assuming $y_{t+1}$ is independent of $x_{1:t}$, the optimization problem in (\ref{eq:objective}) can be reduced to minimizing the cumulative differential entropy, $H(y_{t+1}|z_{1:t}, x_{1:t})$ \cite{atanasov14}. Furthermore, when the belief is Gaussian $B(y_{i,t}) = \mathcal{N}(\hat{y}_{i,t}, \Sigma_{i,t})$, 
\begin{equation}
	H(y_{t}|z_{1:t}, x_{1:t})  = \frac{1}{2} \log\left( (2\pi e)^N \det(\Sigma_t)\right)
\end{equation}
and therefore, the optimization problem becomes :
\begin{equation}
	\text{minimize} \sum_{t=1,\cdots,T} \log \det (\Sigma_{t})
    \label{eq:objective_reduced}
 \end{equation}
We formulate the procedure of the problem in discrete-time as follows. At each step $t$, the robot at the state $x_t$ chooses a control input, $u_t$, based on the prediction on the targets, $\bar{B}(y_{i,t+1})$, to maximize the objective (\ref{eq:objective}). At the same time, the target states will evolve to the next step ($y_t \rightarrow y_{t+1}$). Then, the robot receives measurements, $z_{t+1}$, from the sensor. If some targets are observed, the corresponding belief distributions are updated with the new measurements. This process is repeated at every step. The summary of these steps is presented in Algorithm \ref{table:alg}. Note that we do not study mapping and localization in this paper and assume that the map and the exact odometry of the robot are known to the robot. 

\subsection{Active Target Tracking Network}
An Markov decision process is described by the tuple, $M=<\mathcal{S}, \mathcal{A}, P, R, \gamma>$. $\mathcal{S}$ and $\mathcal{A}$ are state and action spaces, respectively,  $P: \mathcal{S} \times \mathcal{A} \times \mathcal{S} \rightarrow [0,1]$ is a transition probability function, $R: \mathcal{S} \times \mathcal{A} \rightarrow \mathbb{R}$ is a reward function, and $\gamma \in [0,1)$ is a discount factor.  A policy, $\pi$, selects an action that maximizes the expected sum of discounted future rewards. This objective is formally defined as value or value function, $V^{\pi}(s) = \E_{\pi}[\sum^{\infty}_{t=0}\gamma^t R(s_t,a_t)|s_0=s]$, and action-value or Q-value, $Q^{\pi}(s,a) = \E_{\pi}[\sum^{\infty}_{t=0}\gamma^t R(s_t,a_t)|s_0=s, a_0=a]$. 
In the active target tracking problem, we aim to find an optimal policy $\pi^*$ that maximizes the cumulative mutual information in (\ref{eq:objective}), or specifically $-\sum_{t=1,\cdots,T} \log \det (\Sigma_{t})$ in our problem setup. By defining $r_{t+1} = R(s_t, a_t) = -\log \det (\Sigma_{t+1})$, we approximate the objective as a discounted sum. In other words, the value function is :
\begin{equation}
	V^{\pi}(s) = -\E_{\pi}[\sum^{T-1}_{t=0}\gamma^t \log \det (\Sigma_{t+1})|s_0=s]
\end{equation}
The RL action is defined as a control input to the robot. In this paper, we use a finite set of motion primitives for available control inputs. 

Numerous papers have pointed out the limitation of RL methods when goals shift. In general, this requires a complete re-training even though the underlying environment remains the same. However, in an active target tracking setting, not only is a goal position changed in every episode, but a goal position can be continuously changed over time because it is dependent on the uncertainty of beliefs as well as target motions. Since the goal should be implicitly determined by the policy, simple solutions such as including a goal as a part of the RL state \cite{zhu17} will not work. Therefore, we use target information in the agent's perspective for the RL state and reduce the dependency on the training environment.

Information about the target states is essential for the agent to make an optimal decision. The exact target states, however, are unknown to the agent, resulting in the problem being formulated as a partially observable Markov decision process (POMDP) \cite{kaelbling}. Unfortunately, it is known that finding an optimal solution for a general POMDP is intractable \cite{sutton2018reinforcement, papadim}. Instead, we explicitly include the parameters of belief distributions on the targets in the RL state and solve it as an MDP. The target states evolve with their own dynamics independently from the agent's actions. Thus, the agent's decision at time $t$ should be based on predicted target states for $t+1$ rather than the current target states. We define a non-geographical feature vector $\phi_t=[\phi_{1,t}^T, \cdots, \phi_{N,t}^T, (o_t^{(x_t)})^T]^T$.  $\phi_{i,t}$ is composed of the predicted belief state for the $i$-th target in the agent's current frame, its covariance, and the observability of the true $i$-th target : 
\begin{equation}
    \phi_{i,t} \equiv [(\hat{y}_{t+1|t}^{(x_t)})^T, \log \det \Sigma_{i,t+1|t},  \mathbb{I}(y_{i,t} \in \mathcal{O}(x_t))]^T
\end{equation}
$\mathcal{O}(x)$ is an observable space from the robot state $x$ and $\mathbb{I}(\cdot)$ is a boolean function which returns 1 if its given statement is true and 0 otherwise. $o_t^{(x_t)}$ is a coordinate of the closest obstacle point to the agent in the agent frame. 

However, these non-geographical features are not enough for the agent to make an informed decision under all circumstances. As pointed out in the previous sections, the ability to navigate around obstacles is crucial. To achieve this, we need more information than just the closest obstacle point. For example, suppose the agent is located in front of an obstacle and its left corner is closer to the agent than the right corner. It is impossible for the agent to figure out the shorter path to pass the obstacle only with the closest obstacle point. Another important capability of the agent is to explore the current domain when a target is not near the corresponding belief. Note that this exploration means exploring a given environment with a deterministic policy and differs from the exploration problem in RL (action exploration) during learning. To learn to explore in the MDP setting, we build a visit frequency map similar to the occupancy grid mapping. Suppose that $\lambda_c$ is a visit-frequency value for a cell $c$. At $t$, 
\begin{equation}
	\lambda_{c,t} = \begin{cases} 1 & \quad \text{if c is scanned} \\
       					 \lambda_{c,t-1}\cdot c \exp \left(\frac{\bar{v}_t \tau}{r_{sensor}} \right) & \quad \text{otherwise} \end{cases}
           \label{eq:visit_freq}
\end{equation}
where $\bar{v}_t$ is the average speed of the targets from the beliefs, $\tau$ is a sampling period, $r_{sensor}$ is a sensing range of the sensor, and $c$ is a constant factor. Therefore, the most recently visited cells have the value 1.0 and it decays over time as a function of the current target speed estimate.

To reduce the dependency of a learned policy on training environments, we use egocentric maps of the surrounding areas of the current agent position as inputs to the convolutional neural network (CNN) in our architecture. The flatten output of the CNN is concatenated by the non-geographic features $\phi_t$, and then fed to the fully connected network.  Fig. \ref{fig:architecture} shows the illustration of the network architecture and the inputs.

\begin{algorithm}[tb!]
\caption{Active Target Tracking Network (ATTN)}
\begin{algorithmic}[1]
\small
\State Randomly initialize a train Q-network, $Q(s,a|\xi)$
\State Initialize a target Q-network, $Q(s,a|\xi')$ with weights $\xi' \leftarrow \xi$
\State Initialize a replay buffer $D$
\For{episode =1:M}
\State Randomly initialize $x_0$, $y_{i,0}$, $\hat{y}_{i,0}$, $\Sigma_{i,0}$ for $i=1, \cdots, N$
\State Predict $\bar{B}(y_1)$
\State Initialize the RL state: $s_0=f_s(x_0, \bar{B}(y_{1}); M)$)
\For{step $t=0:T-1$}
    \State Choose an action $a_t \sim \pi(s_t)$
    \State Update the agent state $x_{t+1}=f(x_t, a_t)$
    \For{$i=1:N$}
        \State Update the $i$-th target state $y_{i,t+1}=g(y_{i,t})$
        \State Observe the $i$-th target $z_{i,t+1}=h(x_{t+1}, y_{i,t+1})$
        \If{$z_{i,t+1}$ exists}
            \State $B(y_{t+1}) = Z^{-1} p(z_{t+1}|x_{t+1}, y_{t+1}) \bar{B}(y_{t+1})$
        \EndIf
     \EndFor
    \State Compute a reward $r_{t+1} = R(B(y_{t+1}))$
    \State Predict target states $\bar{B}(y_{t+2})$
    \State Update the RL state: $s_{t+1}=f_s(x_{t+1}, \bar{B}(y_{t+2}) ; M)$
    \State $D \leftarrow D \cup \{<s_t, a_t, r_{t+1}, s_{t+1}>\}$
    \State Train the Q-network
\EndFor
\EndFor
\end{algorithmic} \label{table:alg}
\end{algorithm}

\section{Target Tracking Environment} \label{sec:ttenv}
In this section, we describe details of the target tracking environment. It is designed for reinforcement learning practice and follows the OpenAI Gym structure, a popular benchmark simulation environment for the RL community \cite{openaigym}. The source codes can be found at \url{https://github.com/coco66/ttenv}. \textit{Notations:} The xy-position and the orientation in $SE(2)$ are represented with a subscript 1, 2, $\theta$, for example, $(x_1, x_2)\in \mathbb{R}^2$ and $x_{\theta}\in SO(2)$, respectively.

\subsection{Target Model}
\begin{figure}[tb!]
\centering
	\includegraphics[width=0.9\columnwidth]{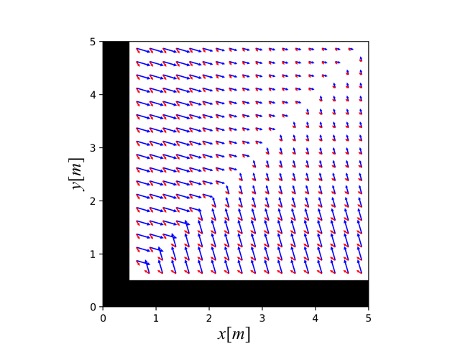}
	\caption{An example of the effect of $\zeta(\cdot)$ in (\ref{eq:nonlinear_target}) at different target positions in the grid. Each arrow corresponds to the velocity components of $\zeta$, $[a_t\tau\cos(\theta_{rot, t}), a_t\tau\sin(\theta_{rot,t})]^T$ when $\tau=0.5, v_{max}=3.0, r_{margin}=0.1, r_{min}=1.0$. The blue arrows are for $v_t=3.0$ and the red arrows are for $v_t=1.0$. The black blocks are obstacles and the target orientation are $-3\pi/4$ [rad] for all.}
	\label{fig:obs_maneuver}
\end{figure}

We design a non-linear target model based on the double integrator with Gaussian noise. In order to maneuver smoothly around obstacles, we add a non-linear term, $\zeta(\cdot)$, that pushes the target away from a nearby obstacle. 
\begin{equation}
    y_{t+1} = Ay_{t} + w_{t} + \zeta(y_{t};M) \qquad w_{t} \sim \mathcal{N}(0,W(q))
    \label{eq:target_model}
\end{equation}
where $y_{t}=[y_{1,t}, y_{2,t}, \dot{y}_{1,t}, \dot{y}_{2,t}]^T$ and  
\[
A=\begin{bmatrix} I_2 &\tau I_2 \\
					0 & I_2  \end{bmatrix}, \qquad W(q)=q \begin{bmatrix} \tau^3/3 I_2 & \tau^2/2 I_2 \\ \tau^2/2 I_2 & \tau I_2 \end{bmatrix}
\]
$q$ is a noise constant and $I_n$ is an $n\times n$ identity matrix. $\zeta(\cdot)$ is a function of the current target state and the map, $M$, and it directs the target away from its closest obstacle which polar coordinate with respect to the current target frame is denoted as $r_o^{(y_t)}, \theta_o^{(y_t)}$.
\begin{equation}
    \zeta(y_{t};M) = [0, 0, a_t\tau\cos(\theta_{rot, t}), a_t\tau\sin(\theta_{rot,t})]^T
    \label{eq:nonlinear_target}
\end{equation}
where
\begin{align}
    a_t =& \frac{\nu_{t}\cos_+ (\theta_o^{(y_t)})}{\max(r_{\min}, r_o^{(y_t)}-r_\text{margin})^2} \\
    \theta_{rot,t} =& \begin{cases} y_{\theta, t} +\theta_o^{(y_t)} - \theta_{rep,t}\quad &\text{for }  \theta_o^{(y_t)} \geq 0 \\[6pt] y_{\theta, t}  + \theta_o^{(y_t)} + \theta_{rep,t} \quad & \text{otherwise} \end{cases} \\
    \theta_{rep,t} =& \frac{\pi}{2}\left( 1 + \frac{1}{1+e^{-(\nu_{t}-\nu_{\max}/2)}} \right)\\ \nu_{t}=&\sqrt{ \dot{y}_{1,t}^2 +  \dot{y}_{2,t}^2}
\end{align}
$r_{min}$ and $r_{margin}$ are design parameters that determine the maximum value of $a_t$ and a margin from an obstacle point with the maximum effect, respectively. $\nu_{max}$ is the maximum target speed, and $\cos_+(x) = \cos(x)$ if $-\pi/2 \leq x \leq \pi/2$ and $0$ otherwise. Fig. \ref{fig:obs_maneuver} illustrates the effect of $\zeta(\cdot)$ around the obstacle in different positions. 
Additionally, if $y_{t+1}$ in (\ref{eq:target_model}) causes a collision with an obstacle, its velocity is changed while its position remains same :
\begin{equation}
    y_{t+1} = y_{t} + (\nu_t+n_{t})[0, 0, \cos(\theta_{rot,t}), \sin(\theta_{rot,t})]^T 
\end{equation}
where $n_{t} \sim \mathcal{N}(0,1)$. 

\subsection{Agent and Observation Models}
The agent follows the differential drive dynamics,
\begin{align}
\begin{bmatrix}
x_{1,t+1} \\ x_{2,t+1} \\ x_{\theta, t+1}
\end{bmatrix} = 
\begin{bmatrix}
x_{1,t} \\x_{2,t} \\ x_{\theta, t}
\end{bmatrix} + 
\begin{bmatrix}
\nu \tau \sinc(\frac{\omega \tau}{2}) \cos (x_{\theta, t} + \frac{\omega \tau}{2})\\
\nu \tau \sinc(\frac{\omega \tau}{2}) \sin (x_{\theta, t} + \frac{\omega \tau}{2})\\
\tau \omega
\end{bmatrix}
\end{align}
controlled by linear and angular velocity commands, $v$ and $\omega$, respectively.

We use a range-bearing sensor which is commonly used in practice in robotics and assume that the agent can uniquely identify different targets.
The observation model of the sensor for each target is:
\begin{equation}
    z_{i,t} = h(x_t, y_{i,t}) + v_t, \hspace{3mm} v_t \sim \mathcal{N}\begin{pmatrix}0, V \end{pmatrix}
    \label{eq:observation_model}
\end{equation}
where $V$ is a observation noise covariance matrix and
\begin{equation*}
    h(x,y) = \begin{bmatrix}r_{x,y}\\\alpha_{x,y}\end{bmatrix} := \begin{bmatrix} \sqrt{(y_1 - x_1)^2 + (y_2 -x_2)^2} \\ \tan^{-1} ((y_2 - x_2)(y_1 - x_1)) - x_{\theta} \end{bmatrix}
\end{equation*}

\subsection{Belief Update}
We use a Kalman Filter to update the beliefs on the targets. While previous works assume that the target model is known to the agent \cite{atanasov14, schlotfeldt18}, we release the assumption by allowing only the partial knowledge, a double integrator with $A$ and $W(q_b)$ in (\ref{eq:target_model}) where $q_b \neq q$, and excluding the $\zeta(\cdot)$ term. In a domain with obstacles, the effect of $\zeta(\cdot)$ is significant, and the belief update becomes considerably inaccurate leading to a more challenging task.

The observation model (\ref{eq:observation_model}) is approximated to a linear model in the Kalman filter and the Jacobian matrix of $h(y, x)$ with respect to $y$ is :
\begin{equation*}
\nabla_y h(x,y) = \frac{1}{r_{x,y}} 
\begin{bmatrix} (y_1 - x_1) & (y_2 - x_2) & \mathbf{0}_{1x2} \\
-\sin (x_{\theta}+\alpha_{x,y}) & \cos (x_{\theta}+\alpha_{x,y})& \mathbf{0}_{1x2} \\ \end{bmatrix}
\end{equation*}


 
\section{Experiments} \label{sec:experiments}
Unlike many benchmark simulation environments in RL, the target tracking environment is highly stochastic and contains considerable randomness. First, the agent, beliefs, and targets can be randomly initialized within a large range in a domain. As discussed further in the following section, different abilities of an agent can be emphasized depending on initialization. Moreover, the noise components in the target motion, the belief model, and the observation model provide additional randomness in performance. Lastly, the agent can discover a target by chance as both the targets and the agent are dynamic and as the target motion is independent of an agent path. Therefore, it is important to have a learning environment that provides diverse samples while to carefully design evaluation environments that reduces a large performance variance across different trials.

We perform experiments in single-target scenarios as well as two-target scenarios. 
When there is more than one target, the maximum velocity of the agent should be much higher than that of targets in order to continuously gather information about the targets while traveling among them. If targets are too far apart, committing to track nearby targets can return higher mutual information in a given time than attempting to track all targets. Therefore, we use different training and evaluation settings for single-target and two-target environments.

\subsection{Algorithms}
We learn a ATTN policy using Assumed Density Filtering Q-learning, a Bayesian Q-learning method which has shown promising performance in stochastic environments and tasks with a large action space \cite{adfq,jeong19_iros}. It is an off-policy temporal difference learning, and thus, we can have an action policy different from the target policy. For the action policy, we use Thompson sampling \cite{thompson} which samples Q-values for all actions given a state from the belief distributions and select an action with the maximum sample value. We use $\gamma=0.99$, the discount factor, which results in the length of the effective horizon as $T_{eff}=687$ ($T_{eff} = \argmax_t \gamma^{t}$ s.t. $\gamma^t > 0.001$). 
The dimension of the egocentric map input is $25 \times 25 \times 5$, and the two convolutional layers consist of 20, $4 \times 4$ filter, with a stride of 3 and 40, $3 \times 3$ filter, with a stride of 2. The following fully connected layers consist of 3 layers with 512 hidden units. The network is trained for 30K steps (or 300 episodes) with three different random seeds (thus, three models).

We compare the ATTN policy with the Anytime Reduced Value Iteration (ARVI) algorithm, an open-source search-based target tracking algorithm that has been verified in both simulations and real robot experiments \cite{schlotfeldt18}. Details of the algorithm is explained in Section \ref{sec:related_work}. The algorithm optimizes the same mutual information objective of ATTN in (\ref{eq:objective_reduced}). It also uses the same observation and agent models and the Kalman filter described in Section \ref{sec:ttenv}. 
ARVI finds a near-optimal path at each step given a specified amount of time denoted as $T_{ARVI}$. Therefore, this allocated planning time is usually equal to or less than the sampling period or a time interval, $\tau$. In this experiment, we set this value to be $T_{ARVI}=\tau=0.5$[s]. The planning horizon for ARVI is set to 12 steps following the original paper. Since the algorithm finds a path in a given time, a longer planning horizon sometimes helps to find a better path but it can also lead to a worse plan since the search space becomes too large. We tested with 1, 5, and 10 for the number of control steps to be executed from a selected path, and 5 showed the best result. 
\subsection{Training Setup}
\begin{figure}[b!]
    \centering
    \includegraphics[width=\columnwidth]{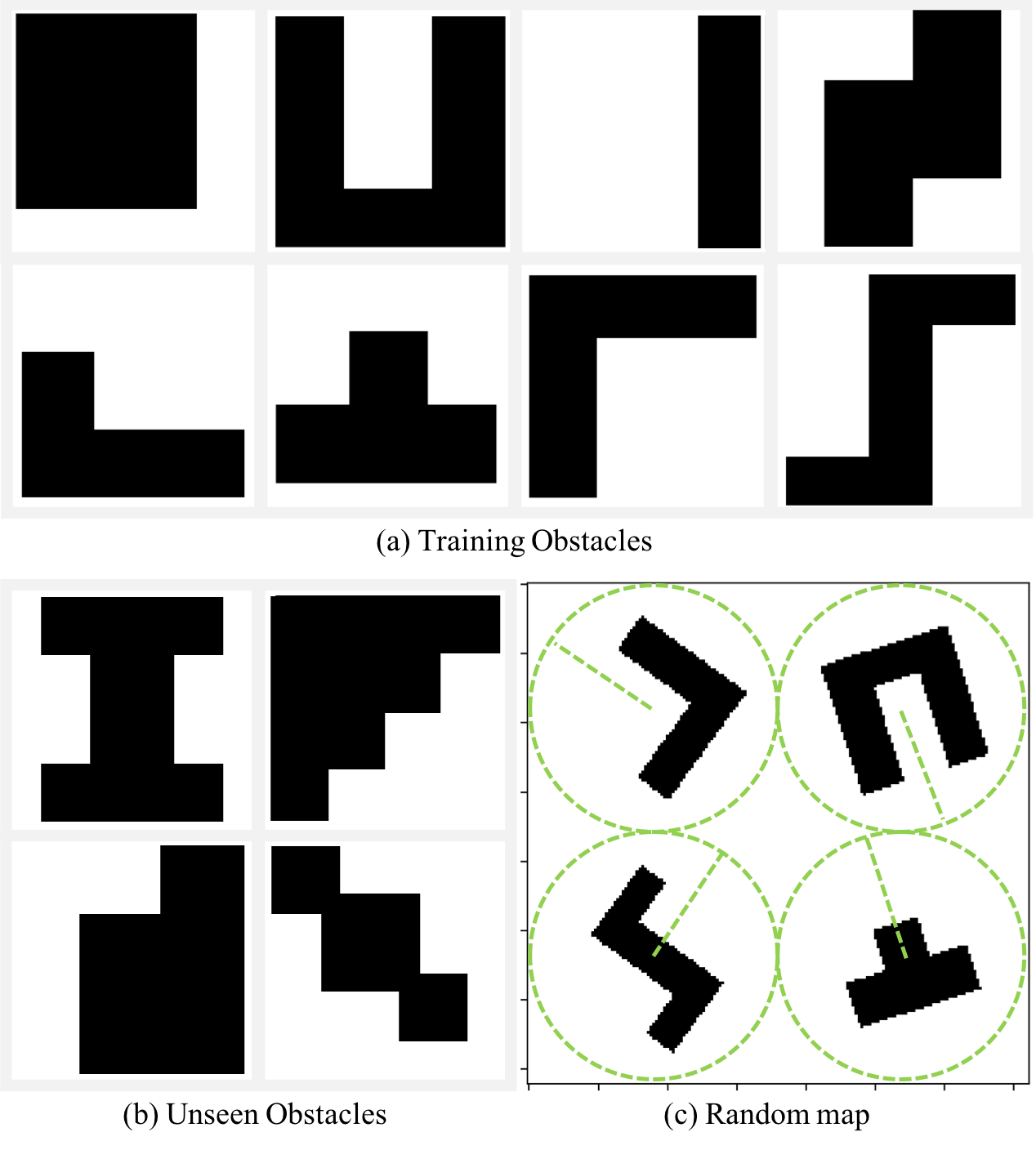}
    \caption{Obstacle polygons for generating a random map. (a) Obstacles used during training, (b) Unseen obstacles, (c) An example map of a randomly generated map.}
    \label{fig:obstacles}
\end{figure}
The training environment is designed so that the learning agent can be exposed to various situations. The system models described in the previous section are used with the following parameters : $\tau=0.5$, $V=\text{diag}(0.2, 0.01)$, $r_\text{margin}=1.0$, $\nu_0=0.0$ [m/s]. The maximum sensing range is $r_\text{sensor}=10$ [m] and its field of view is 120 degrees.  A set of motion primitives, or the action space, is $\mathcal{A}=\{(v,\omega)|v\in\{0,1,2,3\} [m/s], \omega \in\{0, -\pi/2, \pi/2\} [rad/s]\}$ and the time horizon of an episode, $T$ is set to 100.
In single-target scenarios, the noise constants of the target model and the belief model are set as $q=q_b=0.5$, and the maximum target velocity is set as $\nu_{\max}=3.5$ [m/s]. For two-target scenarios, lower values are used for training: $q=q_b=0.2$ and $\nu_{\max}=1.0$ [m/s]. As mentioned above, it is infeasible for an agent to keep tracking multiple targets that are diverging from each other with limited dynamic constraints. Additionally, high noise constant of the target model makes a target to be quickly diverged from its corresponding belief while the agent is tracking another target (when it is not available to cover both targets at once). While this requires an ability to explore near the belief when losing the target, it may also result in a case where committing to one target gives a higher return.

\textbf{Map.} To learn a policy that is robust to various environmental configurations, we randomly generate a map from a set of obstacle candidates as shown in Fig. \ref{fig:obstacles} (a). These obstacles include both convex concave polygons resulting in more challenging navigation tasks. At each episode, four randomly selected obstacles are placed at the center of each quadrant of an empty domain with random orientations (see Fig. \ref{fig:obstacles}(c)). The map resolution (cell size) is 0.4 [m] and the map dimension is $72.4 \times 72.4$ [$\text{m}^2$].

\textbf{Initialization.} For each episode, the robot is randomly initialized in a given domain. In single-target domains, the initial belief position is randomly chosen within a distance between 5.0 to 20.0 [m] from the agent, and the initial target position is randomly placed within (0.0, 20.0)[m] range from the belief. Since the agent may be required to travel between targets in two-target domains, smaller ranges are used -- (5.0, 10.0)[m] range for a distance between beliefs and the agent, and (0.0, 10.0)[m] range for a distance between a belief and its corresponding target. These ranges are chosen considering the maximum sensing range of the agent. Note that having a target placed too far from its corresponding belief can lead to learning a policy that does not trust the belief.

We add a penalty to the reward function when the agent chooses an action that immediately leads to a position within $r_{margin}$ distance to an obstacle. The penalty accelerated the learning by a marginal amount, but did not make any noticeable difference in performance. 
\begin{figure}[b!]
    \centering
    \includegraphics[width=\columnwidth]{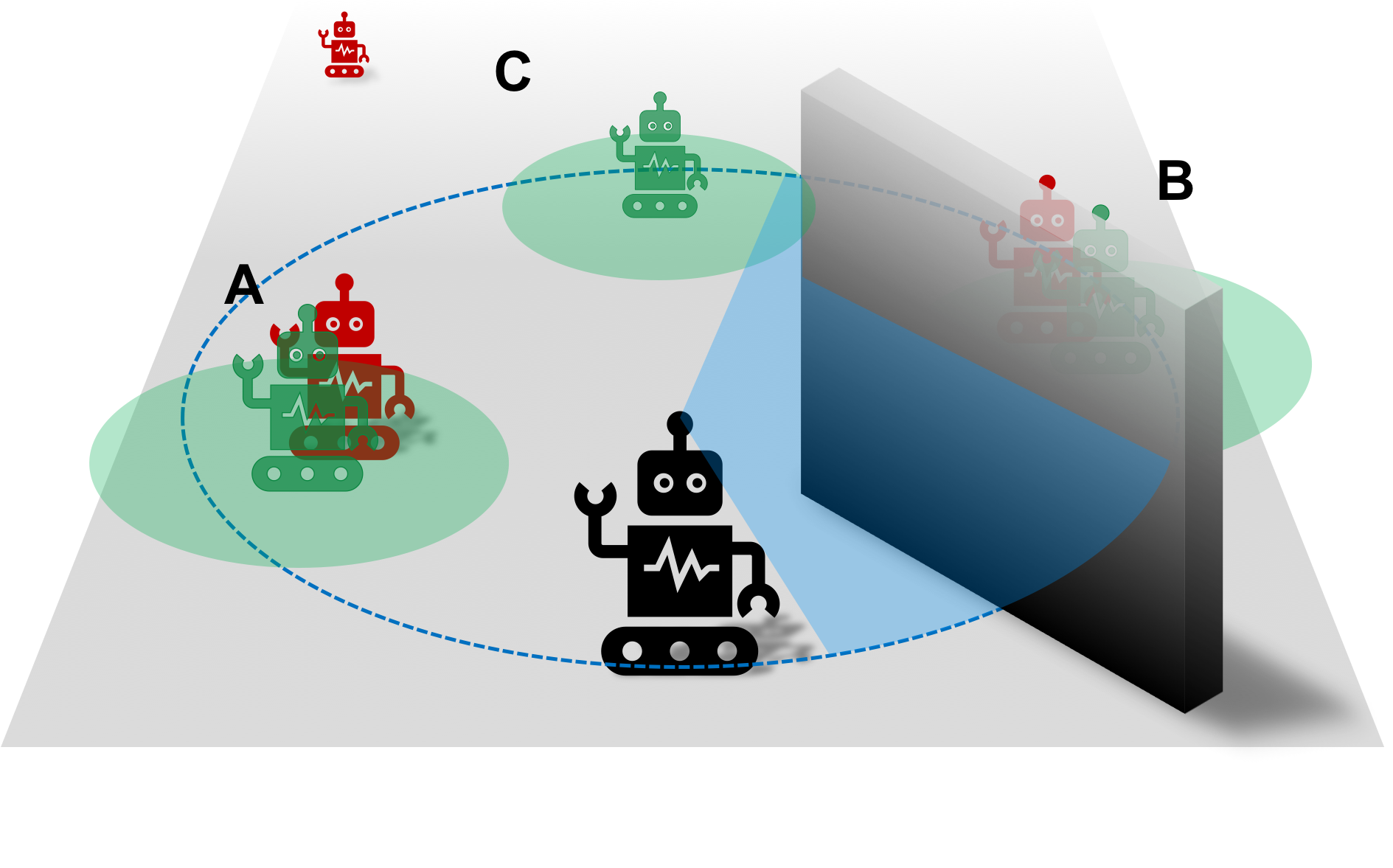}
     \caption{Illustration of the three initialization configurations. The black figure corresponds to the robot with a range-bearing sensor on top. The blue dotted circle is the sensing radius and the faded blue sector indicates a covered area by the sensor. Targets are represented with the red figures. The green figures are beliefs with uncertainty represented as the faded green circles. The black cuboid is an obstacle occluding the target and the belief in the configuration B.}
    \label{fig:init_zone}
\end{figure}
\subsection{Evaluation Setup for Single-Target Domains}
In single-target domains, different sets of initial positions of the robot, targets, and beliefs emphasize different abilities of an agent. For example,
\begin{itemize}
   \item Occlusion of a target and/or its corresponding belief: This requires an ability to navigate around obstacles.
    \item The initial distance between $y_{i,t}$ and $x_t$: A larger value requires the ability to navigate with a long path.
      \item The initial distance between $y_{i,t}$ and $\hat{y}_{i,t}$: A larger value requires the ability to explore the current domain until reaching the target.
\end{itemize}
Therefore, we consider three different initialization setups to evaluate capabilities of the testing algorithms in the sub-tasks of active target tracking -- \textit{in-sight tracking}, \textit{navigation}, \textit{discovery} -- separately.
\begin{enumerate}
\item In-sight tracking task: A target and the corresponding belief are randomly initialized within the sensing range ($||y_{i,t} - x_t||_2 \in [3,10]$), and they are located close to each other ($||y_{i,t} - \hat{y}_{i,t}||_2 \in [0,3]$). The case A in Fig. \ref{fig:init_zone} illustrates this initial condition. We experiment with different values for  $q$ and $\nu_\text{max}$. 
\item Navigation task: Both a target and the corresponding belief are initialized relatively far from the agent ($||y_{i,t} - x_t||_2 \in [15,20]$), and occluded as described in Fig. \ref{fig:init_zone} B. They are located close to each other ($||y_{i,t} - \hat{y}_{i,t}||_2 \in [0,3]$).  
\item Discovery task : The initial belief is located within the sensing range ($||\hat{y}_{i,t} - x_t||_2 \in [3,10]$), but the target is initialized far from the belief  ($||y_{i,t} - \hat{y}_{i,t}||_2 \in [15,20]$) as illustrated in Fig. \ref{fig:init_zone} C. Thus, the agent won't be able to discover the target by simply reaching the belief location and scanning around.
\end{enumerate}

We first generated a set of 10 episodes for each evaluation task. In each episode, the target trajectory, initial robot and belief states, and map configuration are randomly generated and they differ across episodes.
Three trained models of ATTN with different random seeds are evaluated on the evaluation sets. ARVI is also evaluated on the evaluation sets with three random seeds. To prevent a case where the target approaches the robot and is found by luck, the target starts to move once it is observed by the robot for the first time in each episode.
\begin{figure}[b!]
    \centering
    \includegraphics[width=\columnwidth]{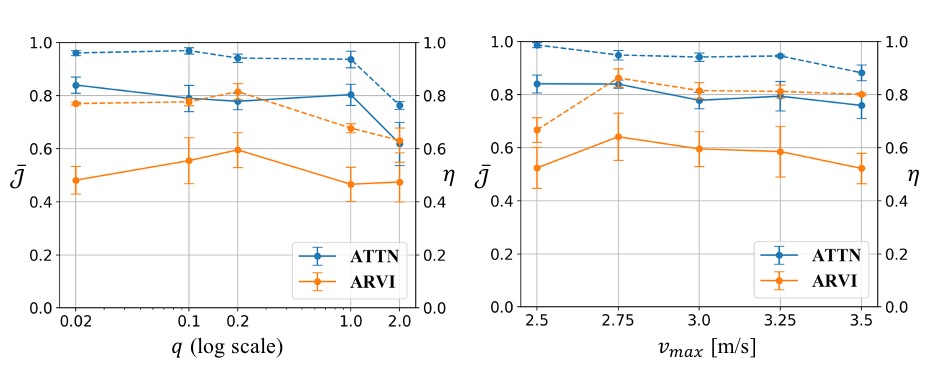}\\
    \caption{$\bar{J}$ (solid line) and $\eta$ (dotted line) of ATTN (blue) and ARVI (yellow) in environments with different $q$ and $\nu_{\max}$ values. The error bars represent the standard deviation across 10 episodes. The mean values are averaged values over different seeds and 10 episodes. Left: $\nu_{\max}=3.0$[m/s] and $q\in\{0.02, 0.1, 0.2, 1.0, 2.0\}$. Right: $q=0.2$, $\nu_{max}\in \{2.5, 2.75, 3.0, 3.25, 3.5\}$}
    \label{fig:eval_qv}
\end{figure}

\subsection{Results in Single-Target Domains}
\begin{figure}[tb!]
    \centering
    \includegraphics[width=\columnwidth]{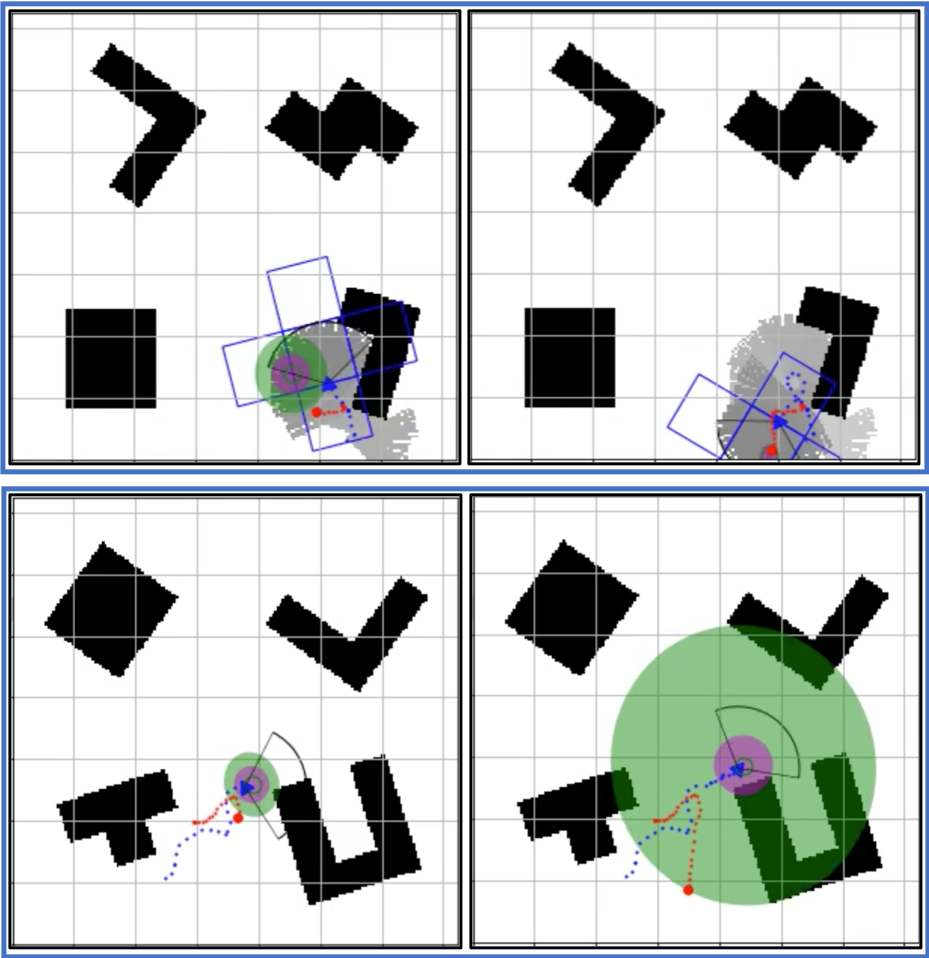}\\
    \caption{Examples of ATTN (top) and ARVI (bottom) when the robot loses the target. The left figures are a few steps after the robot loses the target, and the right figures are after 10 steps passed. The blue triangle is the robot and the red circle is the current target position. The blue and red dots are paths of the robot and target so far. The green circle indicates the belief position. The green and purple shaded circles represent the position and velocity uncertainty of the belief, respectively. The circular sector is the sensing area. In the top figures, visited cells are filled with gray color based on $\lambda_{c,t}$ in (\ref{eq:visit_freq}), and the five blue squares indicate areas in the local map input.}
    \label{fig:recovery}
\end{figure}
First, the robustness of the ATTN policy and ARVI is evaluated in the in-sight tracking task. In particular, we are interested in tracking fast and anomalous targets. Thus, we test ATTN and ARVI with different values for $q$ in the target model (\ref{eq:target_model}) and $\nu_{\max}$. The larger the $q$ value, the more deviated the belief state is from the target state. Furthermore, $q$ affects how much the target speed evolves over time. If the $q$ value is small, the target may never reach the maximum target speed in a given time horizon. $q_b$ is set to be 0.5 as same as the value used in training.

To quantitatively measure the performance of the algorithms, we consider mutual information, or in particular, a normalized sum of  negative log of determinant of covariance in predictions within an episode:
\begin{equation}
\bar{J} = \frac{-\sum_{t=1,\cdots,T} \log \det (\Sigma_{t+1|t}) - J_{\min}}{J_{\max} - J_{\min}} \in [0,1]
\end{equation}The lower bound, $J_{\min}$, is found when there is no observation in an episode, and the belief is updated only by the prediction step in the Kalman filter. The upper bound, $J_{\max}$, is met when the target is observed at every step. According to Theorem 4 in \cite{sinopoli04},  $J_{\max} = - T\log\det(W)$ and $J_{\min} = - \sum_{t=1,\cdots,T} \log\det(\Sigma_{t+1})$ for $\Sigma_{t+1} = A\Sigma_t A^T + W$. We additionally evaluate resilience, $\eta \in [0,1]$, defined as the number of times the target is re-discovered by the sensor divided by the number of times the target is lost.

\begin{figure}[t!]
    \centering
    \includegraphics[width=\columnwidth]{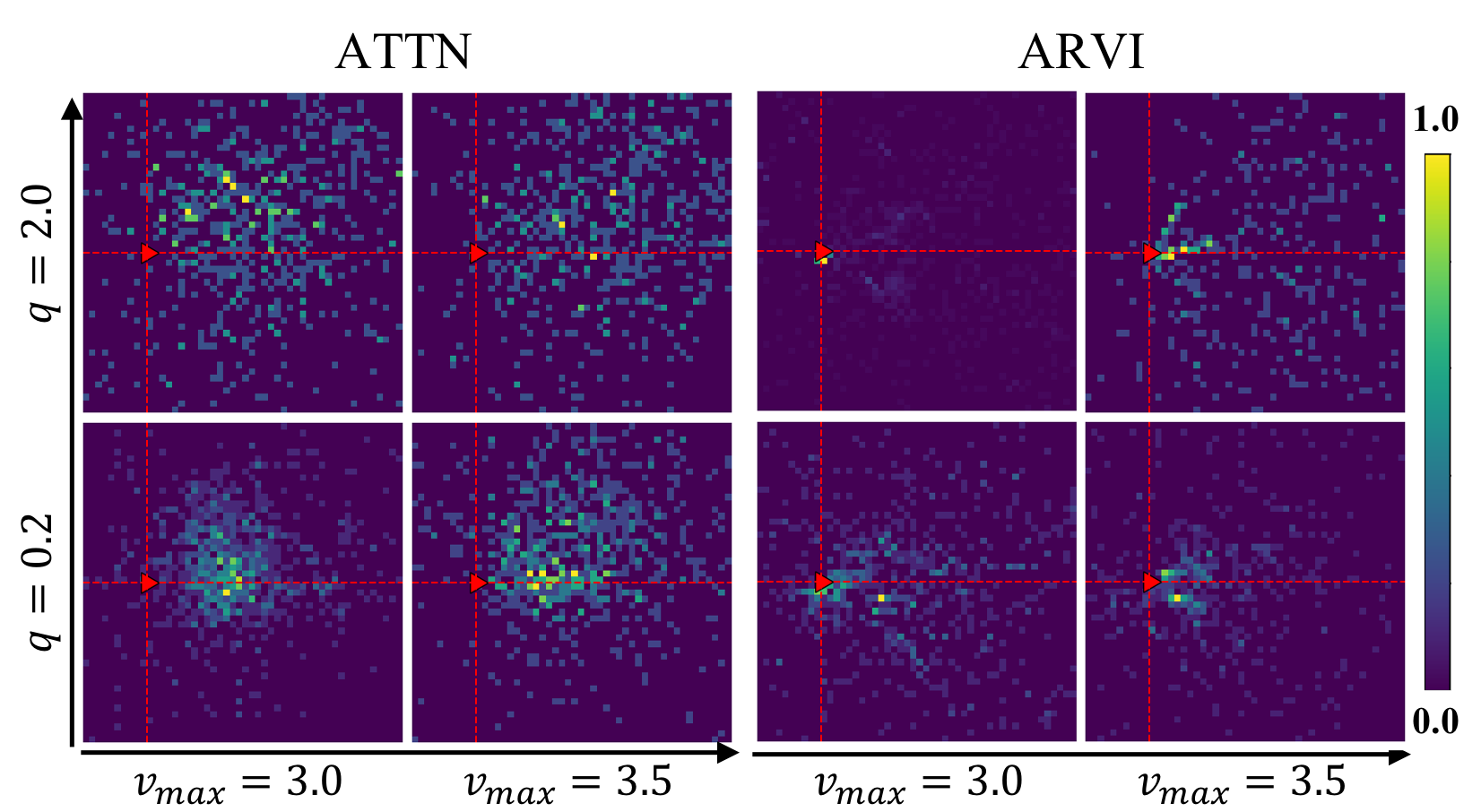}
    \caption{Density plots of belief positions in the agent frame during 10 episodes with different values for $q$ and $\nu_{max}$. The red triangle is the robot position (0.0, 0.0) and the horizontal and vertical red dotted lines are x and y axis of the agent frame, respectively. $x\in (-2.0,8.0)$ and $y\in(-5.0, 5.0)$[m].}
    \label{fig:rel_target_density}
\end{figure}

The left plot in Fig. \ref{fig:eval_qv} shows the average performances of both algorithms for $q\in \{0.02, 0.1, 0.2, 1.0, 2.0\}$ and $\nu_{\max}=3.0$, and the right plot shows their average performances for $q=0.2$ and $\nu_{\max}\in\{2.5, 2.75, 3.0, 3.25, 3.5\}$. Note that the maximum robot linear speed is 3.0 [m/s], and thus, it is likely that the robot will often lose the target when $\nu_{\max}=3.25$ and $\nu_{\max}=3.5$. In both $\bar{J}$ and $\eta$, ATTN outperformed ARVI. As expected, $\bar{J}$ and $\eta$ decrease as the target motion becomes noisier. Surprisingly, $\nu_{\max}$ did not have a significant impact on either $\bar{J}$ or $\eta$. This shows that the robot is capable of tracking a target that is faster than itself. The resilience tracks similarly with $\bar{J}$ indicating that the resilience is a significant factor of the performance. ATTN learns to explore near the belief when the target is not observed at the belief location and the uncertainty is high. Fig. \ref{fig:recovery} shows an example of each algorithm when the agent loses the target. In the top figures, the belief is located to the left of the robot, but the robot with the ATTN policy chooses to turn right instead to explore the surrounding areas. After a few steps, the robot re-discovers the target. On the other hand, ARVI in the bottom figures greedily follows the belief, failing to discover the target even though the uncertainty is relatively high and no observation is received. 
Additionally, ATTN does not track the belief as tightly as ARVI, and instead leaves some buffer space to provide maneuverability when the target quickly changes its direction. Fig. \ref{fig:rel_target_density} is density plots of belief positions with respect to the robot during 10 episodes. The red dotted lines are the x-axis and y-axis of the robot frame, and the x-axis is the robot heading direction. Overall, ATTN results in scattered density plots while the belief positions in the ARVI plots are mostly concentrated in a few small areas.

\begin{figure}[b!]
    \includegraphics[width=\columnwidth]{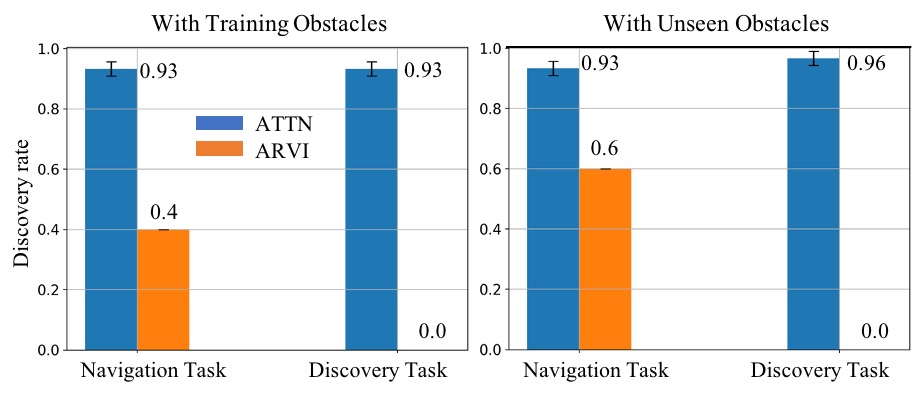}
    \caption{Performance evaluation for \textit{Discovery} and \textit{Navigation} tasks. Left: Evaluated in 10 different environments randomly generated by obstacles used in training. Right: Evaluated in 10 different environments randomly generated by unseen obstacles during training.}
    \label{fig:eval_discovery}
\end{figure}
\begin{figure}[b!]
    \centering
    \includegraphics[width=\columnwidth]{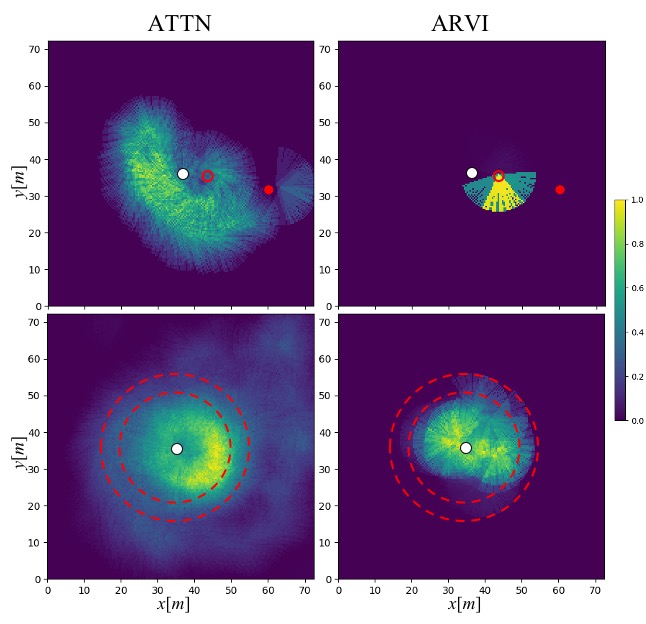}
    \caption{Density maps of scanned areas by the robot's sensor over single episode (top) and 20 episodes (bottom). The white circle are the initial position of the robot. The red filled circles in the top figures are the initial target positions, and the red hollow circles in the top figures are the initial belief positions. In the bottom figures, the target is randomly initialized in the area between the red dotted circles.}
    \label{fig:visit_freq_map}
\end{figure}

For the navigation and discovery tasks, we measure a discovery rate to evaluate the performance of the algorithms. This discovery rate is defined as the number of episodes in which the target is found divided by the number of total episodes (=10). The faster the robot finds the target, the higher the resulting $\bar{J}$. However, $\bar{J}$ can vary significantly depending on the initial positions of the setup, and therefore is a poor measure for these tasks. 

The results of the navigation task are depicted in the left figure of Fig. \ref{fig:eval_discovery}. The ATTN policy finds the target 93\% of the time on average (models trained with three different random seeds). On the other hand ARVI successfully finds the target only 40\% of the time. We observe that ARVI fails to find a path to the target given $T_{ARVI}$ when target is located far from the agent and is occluded by concave polygon obstacles. 

Since ARVI uses a forward simulation, it guarantees to avoid any collision with a perfect map information. While ATTN does not provide the guarantee for the collision avoidance, ATTN results in 0.4 times of collision attempts on average in the navigation task and no collision attempt in the discovery task.

In the discovery task, as defined earlier, the initial target is placed far from the initial belief location requiring the robot to explore to find the target. As shown in Fig. \ref{fig:eval_discovery}, ARVI completely fails to find the target in any of the episodes in this task while ATTN finds it 93\% of the time on average. The four figures in Fig. \ref{fig:visit_freq_map} are density maps of areas that the robot's sensor has scanned in the global map. The top figures are examples of a single episode (left: ATTN, right: ARVI) and the bottom figures are examples of 20 episodes with random initialization. To solely evaluate the exploration capability, no obstacles are present in the map. The figures show that ATTN explores the global domain broadly when the target is not found near the belief while ARVI commits to the incorrect belief and leaves the global domain unexplored. This difference explains why ATTN achieves an 93\% discovery rate while ARVI never finds the target. 
The computation time for planning of ATTN is 0.12 [sec] on average, which is much faster than the allocated computation used for ARVI ($=0.5$ [sec]) while ATTN has a significantly longer planning horizon.  

Additional to the experiments described in the previous sections, we test the learned ATTN policy in environments with unseen obstacles during training (see Fig. \ref{fig:obstacles}). The results in the right figure of Fig. \ref{fig:eval_discovery} shows that the discovery rates of ATTN in both tasks are 93\% and 96\%, similar to the results in the training environments. Likewise, ARVI results in 60\% and 0\% discovery rates. Additionally, ATTN shows 0.4 and 0.0 collision attempts on average for the navigation and discovery tasks, respectively. These results demonstrate that the learned policy performs well in unseen environments promoting the benefit of using local information. Note that ARVI is an online search-based planning algorithm without having a training stage, and therefore it is not affected by a new set of obstacles. 
\begin{figure}[t!]
    \centering
    \includegraphics[width=\columnwidth]{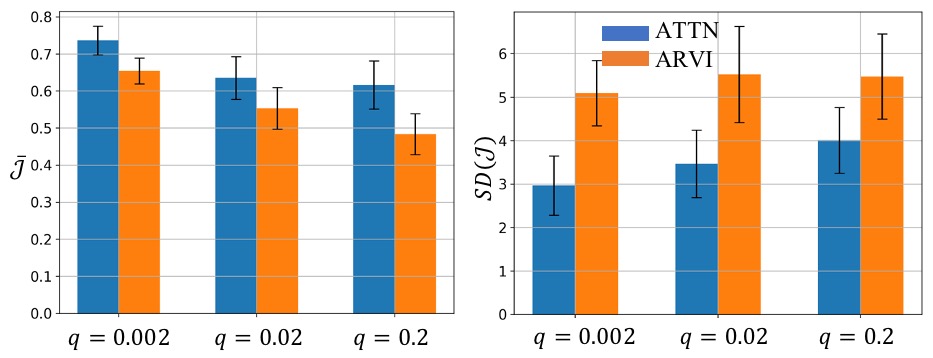}
    \caption{Performance evaluation on $N=2$ targets. Left: The normalized mean of log determinant of belief covariances averaged over 10 episodes and the error bars indicate standard deviations. Right: Standard deviation of log determinant of belief covariances averaged over 10 episodes and the error bars indicate their standard deviations.}
    \label{fig:multi_target_eval}
\end{figure}
\subsection{Results in Two-Target Domains}
\begin{figure}[b!]
    \centering
    \includegraphics[width=0.85\columnwidth]{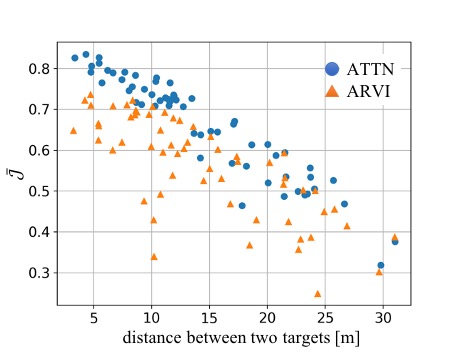}
    \caption{The relation between an average distance between two targets during an episode and $\bar{J}$ (blue circle: ATTN, yellow triangle: ARVI). }
    \label{fig:eval_multi_distance}
\end{figure}
Unlike the single-target domains (where once the agent is nearby the target, the navigation and discovery abilities are not much needed), the agent may need to use all three abilities interchangeably throughout an episode or to use them at the same time. For instance, the agent must navigate from one target to another one when they are far apart. Moreover, while the agent is tracking one target, the other belief diverges from its corresponding target which requires the agent to explore to find the target. Lastly, the agent may explore to discover one target while tracking the other target when they both are around the agent's sensing range. Therefore, instead of evaluating the algorithms in the three subtasks, we evaluate the algorithms with an initialization similar to the in-sight tracking task in the single-target evaluation at different noise constants $q=\{0.002, 0.02, 0.2 \}$ and $v_{\text{max}}=1.0$ [m/s].

The average normalized objective, $\bar{J}$, are presented in the left figure of Fig. \ref{fig:multi_target_eval}. In all $q$ values, ATTN consistently outperforms ARVI. The right figure shows the average standard deviation between $\bar{J}$ of two targets. A lower value for $SD(\bar{J})$ indicates that the agent tracks the targets while balancing the objective brought by both targets. Similar to the single target case, ARVI aggressively tracks the beliefs. It especially harms the performance in a multi-target tracking case since the agent is often required to travel between two targets and a belief diverges from its corresponding target while the agent is tracking another one.

Fig.\ref{fig:eval_multi_distance} shows how the performance of ATTN and ARVI in terms of $\bar{J}$ decreases as a distance between two targets increases. Although ATTN shows a higher performance than ARVI, the value drops in general which indicates the difficulty of the problem. 
\section{Conclusion and Discussion}
In this paper, we present Active Target Tracking Network (ATTN), an RL method for active target tracking that learns a unified policy capable of three main tasks - \textit{In-sight tracking}, \textit{Navigation}, and \textit{Discovery}. To demonstrate, we train an ATTN policy in a target tracking environment described in Section \ref{sec:ttenv} where a mobile robot is tasked with tracking mobile targets using noisy measurements from its onboard range-bearing sensor. The learned ATTN policy shows a robust performance for agile and anomalous target motions, despite the true target model differing from the target model used in the agent’s belief update. Moreover, the policy was able to navigate through obstacles to reach distant targets. Finally, it learns to explore surrounding areas and discovers a target when its belief on the target is inaccurate, while the existing algorithm failed to do so in all test examples. 

\subsection{Learning with Uncertainty}
Although an RL method requires an extended training stage to obtain a policy, it provides flexibility in expanding a problem domain or changing system models. The objective of a problem is implicitly included in a reward function, and an optimal policy is learned without requiring knowledge on models. While we use partial knowledge on a target model and an observation model to update beliefs using a Kalman filter, we are able to use a target model that differs from the model known to the beliefs without violating any assumption or constraint required for an algorithm. Therefore, the target tracking scenarios considered in this article are more challenging in terms of stochasticity, uncertainty, and imperfect prior knowledge, compared to scenarios presented in previous works \cite{chung06,hoffmann,atanasov14,schlotfeldt18}. During learning, the learning agent is exposed to diverse and challenging experiences and is able to \textit{learn to track with uncertain beliefs}, \textit{learn to navigate}, and \textit{learn to explore}. 

\subsection{Stochasticity of Tasks}
Active target tracking is not a trivial task for RL. The states of the learning agent and of the environment including targets and obstacles are high-dimensional, continuous, and stochastic. The recent advancement of deep RL have shown promising results in tasks where an RL state is continuous or high-dimensional -- for example, raw screen image in video games or continuous joint angles \cite{mnih2013, schulman2015trust}. Yet, the deterministic dynamics of popular deep RL benchmark environments such as the Arcade Learning Environment (ALE) and Mujoco has raised concern in the scientific community as the successfully evaluated algorithms can fail when extended to new domains  \cite{machado18}. In most real-world problems, a transition function and/or a reward function are likely to be stochastic, resulting in challenges applying such approaches to more realistic problems. The active target tracking task is highly stochastic, even in its simulation setting. The innate partial observability feature brings high stochasticity in the task. A subsequent RL state is determined not only by the current agent action but also the real yet partially observable or unknown target. Therefore, the same action can result in very different RL states as the belief can be drastically updated due to a new measurement if the corresponding target is around as opposed to no target is observed. It is shown in the previous work \cite{jeong19_iros} that popular benchmark deep Q-learning methods, deep Q-network and double deep Q-network, failed to achieve optimal performance in relatively simple settings. Thus, it is important to note that the environment is stochastic and an algorithm capable in such a highly stochastic domain should be considered. 

\subsection{United Policy}
One of the key contributions of this study is that we expand the problem domain and tackle three major capabilities with one united policy. One might argue that we could achieve similar performance by having three different algorithms for each capability and have an additional algorithm to heuristically decide which one to use at each step. For example, it is possible to improve the performance of ARVI by using an exploration method when a target is not detected. However, this not only burdens the computational cost but loses the interchangeable flexibility among different capabilities.

\subsection{Future Work}
We have pointed out the limitation of a single agent tracking multiple targets. This limitation, in fact, has been one of the major motivations for multi-agent tracking studies \cite{charrow14, motchallenge, schlotfeldt18}. Separately, multi-agent reinforcement learning has been widely studied for various applications including robotics, telecommunications, and distributed control \cite{bu2008comprehensive, omidshafiei2017deep, lanctot2017unified}. Combining advancements from these two fields of literature would present an interesting and more practical result in scalable target tracking problems.

Additionally, this work can be extended to adversarial target tracking problems where targets try not to be seen by the agent. By training an intelligent target that adapts its behavior to avoid being tracked by the agent with its newest policy, the agent can continuously improve the policy according to the updated target behavior. Such adversarial learning or self-supervised reinforcement learning approaches to active target tracking would be an interesting direction for future studies \cite{goodfellow2014generative, baker2019emergent}.

\bibliographystyle{IEEEtran}
\bibliography{IEEEabrv, aia_bib}

\begin{thebibliography}{10}
\providecommand{\url}[1]{#1}
\csname url@samestyle\endcsname
\providecommand{\newblock}{\relax}
\providecommand{\bibinfo}[2]{#2}
\providecommand{\BIBentrySTDinterwordspacing}{\spaceskip=0pt\relax}
\providecommand{\BIBentryALTinterwordstretchfactor}{4}
\providecommand{\BIBentryALTinterwordspacing}{\spaceskip=\fontdimen2\font plus
\BIBentryALTinterwordstretchfactor\fontdimen3\font minus
  \fontdimen4\font\relax}
\providecommand{\BIBforeignlanguage}[2]{{%
\expandafter\ifx\csname l@#1\endcsname\relax
\typeout{** WARNING: IEEEtran.bst: No hyphenation pattern has been}%
\typeout{** loaded for the language `#1'. Using the pattern for}%
\typeout{** the default language instead.}%
\else
\language=\csname l@#1\endcsname
\fi
#2}}
\providecommand{\BIBdecl}{\relax}
\BIBdecl

\bibitem{rybski00}
P.~E. Rybski, S.~A. Stoeter, M.~D. Erickson, M.~Gini, D.~F. Hougen, and
  N.~Papanikolopoulos, ``A team of robotic agents for surveillance,'' in
  \emph{Proceedings of the fourth international conference on autonomous
  agents}, 2000, pp. 9--16.

\bibitem{hilal13}
A.~Hilal, ``An intelligent sensor management framework for pervasive
  surveillance,'' 2013.

\bibitem{kumar04}
V.~Kumar, D.~Rus, and S.~Singh, ``Robot and sensor networks for first
  responders,'' \emph{IEEE Pervasive computing}, vol.~3, no.~4, pp. 24--33,
  2004.

\bibitem{soatto11}
S.~Soatto, ``Steps towards a theory of visual information: Active perception,
  signal-to-symbol conversion and the interplay between sensing and control,''
  \emph{arXiv preprint arXiv:1110.2053}, 2011.

\bibitem{zhang17}
D.~Zhang, H.~Maei, X.~Wang, and Y.-F. Wang, ``Deep reinforcement learning for
  visual object tracking in videos,'' \emph{arXiv preprint arXiv:1701.08936},
  2017.

\bibitem{luo17}
W.~Luo, P.~Sun, F.~Zhong, W.~Liu, T.~Zhang, and Y.~Wang, ``End-to-end active
  object tracking via reinforcement learning,'' \emph{arXiv preprint
  arXiv:1705.10561}, 2017.

\bibitem{jayaraman18}
D.~Jayaraman and K.~Grauman, ``Learning to look around: Intelligently exploring
  unseen environments for unknown tasks,'' in \emph{Proceedings of the IEEE
  Conference on Computer Vision and Pattern Recognition}, 2018, pp. 1238--1247.

\bibitem{dunbabin12}
M.~Dunbabin and L.~Marques, ``Robots for environmental monitoring: Significant
  advancements and applications,'' \emph{IEEE Robotics \& Automation Magazine},
  vol.~19, no.~1, pp. 24--39, 2012.

\bibitem{choi09}
H.-L. Choi, ``Adaptive sampling and forecasting with mobile sensor networks,''
  Ph.D. dissertation, Massachusetts Institute of Technology, 2009.

\bibitem{chung06}
T.~H. Chung, J.~W. Burdick, and R.~M. Murray, ``A decentralized motion
  coordination strategy for dynamic target tracking,'' in \emph{Proceedings
  2006 IEEE International Conference on Robotics and Automation, 2006. ICRA
  2006.}\hskip 1em plus 0.5em minus 0.4em\relax IEEE, 2006, pp. 2416--2422.

\bibitem{kreucher05}
C.~M. Kreucher, \emph{An information-based approach to sensor resource
  allocation}.\hskip 1em plus 0.5em minus 0.4em\relax University of Michigan,
  2005.

\bibitem{huber09}
M.~Huber, \emph{Probabilistic framework for sensor management}.\hskip 1em plus
  0.5em minus 0.4em\relax KIT Scientific Publishing, 2009, vol.~7.

\bibitem{atanasov14}
N.~Atanasov, J.~Le~Ny, K.~Daniilidis, and G.~J. Pappas, ``Information
  acquisition with sensing robots: Algorithms and error bounds,'' in \emph{2014
  IEEE International Conference on Robotics and Automation (ICRA)}.\hskip 1em
  plus 0.5em minus 0.4em\relax IEEE, 2014, pp. 6447--6454.

\bibitem{schlotfeldt18}
B.~Schlotfeldt, D.~Thakur, N.~Atanasov, V.~Kumar, and G.~J. Pappas, ``Anytime
  planning for decentralized multirobot active information gathering,''
  \emph{IEEE Robotics and Automation Letters}, vol.~3, no.~2, pp. 1025--1032,
  2018.

\bibitem{hollinger2014sampling}
G.~A. Hollinger and G.~S. Sukhatme, ``Sampling-based robotic information
  gathering algorithms,'' \emph{The International Journal of Robotics
  Research}, vol.~33, no.~9, pp. 1271--1287, 2014.

\bibitem{jeong19_iros}
H.~Jeong, B.~Schlotfeldt, H.~Hassani, M.~Morari, D.~D. Lee, and G.~J. Pappas,
  ``Learning q-network for active information acquisition,'' in \emph{2019
  IEEE/RSJ International Conference on Intelligent Robots and Systems
  (IROS)}.\hskip 1em plus 0.5em minus 0.4em\relax IEEE, 2019, pp. 6822--6827.

\bibitem{motchallenge}
L.~Leal-Taix{\'e}, A.~Milan, I.~Reid, S.~Roth, and K.~Schindler, ``Motchallenge
  2015: Towards a benchmark for multi-target tracking,'' \emph{arXiv preprint
  arXiv:1504.01942}, 2015.

\bibitem{giusti15}
A.~Giusti, J.~Guzzi, D.~C. Cire{\c{s}}an, F.-L. He, J.~P. Rodr{\'\i}guez,
  F.~Fontana, M.~Faessler, C.~Forster, J.~Schmidhuber, G.~Di~Caro
  \emph{et~al.}, ``A machine learning approach to visual perception of forest
  trails for mobile robots,'' \emph{IEEE Robotics and Automation Letters},
  vol.~1, no.~2, pp. 661--667, 2015.

\bibitem{whitehead90}
S.~D. Whitehead and D.~H. Ballard, ``Active perception and reinforcement
  learning,'' in \emph{Machine Learning Proceedings 1990}.\hskip 1em plus 0.5em
  minus 0.4em\relax Elsevier, 1990, pp. 179--188.

\bibitem{williams1992simple}
R.~J. Williams, ``Simple statistical gradient-following algorithms for
  connectionist reinforcement learning,'' \emph{Machine learning}, vol.~8, no.
  3-4, pp. 229--256, 1992.

\bibitem{yamauchi1997frontier}
B.~Yamauchi, ``A frontier-based approach for autonomous exploration,'' in
  \emph{Proceedings 1997 IEEE International Symposium on Computational
  Intelligence in Robotics and Automation CIRA'97.'Towards New Computational
  Principles for Robotics and Automation'}.\hskip 1em plus 0.5em minus
  0.4em\relax IEEE, 1997, pp. 146--151.

\bibitem{gonzalez2002navigation}
H.~H. Gonz{\'a}lez-Banos and J.-C. Latombe, ``Navigation strategies for
  exploring indoor environments,'' \emph{The International Journal of Robotics
  Research}, vol.~21, no. 10-11, pp. 829--848, 2002.

\bibitem{amigoni2010information}
F.~Amigoni and V.~Caglioti, ``An information-based exploration strategy for
  environment mapping with mobile robots,'' \emph{Robotics and Autonomous
  Systems}, vol.~58, no.~5, pp. 684--699, 2010.

\bibitem{bourgault}
F.~{Bourgault}, A.~A. {Makarenko}, S.~B. {Williams}, B.~{Grocholsky}, and H.~F.
  {Durrant-Whyte}, ``Information based adaptive robotic exploration,'' in
  \emph{IEEE/RSJ International Conference on Intelligent Robots and Systems},
  vol.~1, 2002, pp. 540--545 vol.1.

\bibitem{kollar2008trajectory}
T.~Kollar and N.~Roy, ``Trajectory optimization using reinforcement learning
  for map exploration,'' \emph{The International Journal of Robotics Research},
  vol.~27, no.~2, pp. 175--196, 2008.

\bibitem{lauri2016planning}
M.~Lauri and R.~Ritala, ``Planning for robotic exploration based on forward
  simulation,'' \emph{Robotics and Autonomous Systems}, vol.~83, pp. 15--31,
  2016.

\bibitem{bagnell04}
J.~A. Bagnell, S.~M. Kakade, J.~G. Schneider, and A.~Y. Ng, ``Policy search by
  dynamic programming,'' in \emph{Advances in neural information processing
  systems}, 2004, pp. 831--838.

\bibitem{gupta17}
S.~Gupta, J.~Davidson, S.~Levine, R.~Sukthankar, and J.~Malik, ``Cognitive
  mapping and planning for visual navigation,'' in \emph{Proceedings of the
  IEEE Conference on Computer Vision and Pattern Recognition}, 2017, pp.
  2616--2625.

\bibitem{kahn17}
G.~Kahn, T.~Zhang, S.~Levine, and P.~Abbeel, ``Plato: Policy learning using
  adaptive trajectory optimization,'' in \emph{2017 IEEE International
  Conference on Robotics and Automation (ICRA)}.\hskip 1em plus 0.5em minus
  0.4em\relax IEEE, 2017, pp. 3342--3349.

\bibitem{zhu17}
Y.~Zhu, R.~Mottaghi, E.~Kolve, J.~J. Lim, A.~Gupta, L.~Fei-Fei, and A.~Farhadi,
  ``Target-driven visual navigation in indoor scenes using deep reinforcement
  learning,'' in \emph{2017 IEEE international conference on robotics and
  automation (ICRA)}.\hskip 1em plus 0.5em minus 0.4em\relax IEEE, 2017, pp.
  3357--3364.

\bibitem{kaelbling}
L.~P. Kaelbling, M.~L. Littman, and A.~R. Cassandra, ``Planning and acting in
  partially observable stochastic domains,'' \emph{Artificial intelligence},
  vol. 101, no. 1-2, pp. 99--134, 1998.

\bibitem{sutton2018reinforcement}
R.~S. Sutton and A.~G. Barto, \emph{Reinforcement learning: An
  introduction}.\hskip 1em plus 0.5em minus 0.4em\relax MIT press, 2018.

\bibitem{papadim}
C.~H. Papadimitriou and J.~N. Tsitsiklis, ``The complexity of markov decision
  processes,'' \emph{Mathematics of operations research}, vol.~12, no.~3, pp.
  441--450, 1987.

\bibitem{openaigym}
G.~Brockman, V.~Cheung, L.~Pettersson, J.~Schneider, J.~Schulman, J.~Tang, and
  W.~Zaremba, ``Openai gym,'' 2016.

\bibitem{adfq}
H.~Jeong, C.~Zhang, G.~J. Pappas, and D.~D. Lee, ``Assumed density filtering
  q-learning,'' in \emph{Proceedings of the 28th International Joint Conference
  on Artificial Intelligence}.\hskip 1em plus 0.5em minus 0.4em\relax AAAI
  Press, 2019, pp. 2607--2613.

\bibitem{thompson}
W.~R. Thompson, ``On the likelihood that one unknown probability exceeds
  another in view of the evidence of two samples,'' \emph{Biometrika}, vol.~25,
  no. 3/4, pp. 285--294, 1933.

\bibitem{sinopoli04}
B.~Sinopoli, L.~Schenato, M.~Franceschetti, K.~Poolla, M.~I. Jordan, and S.~S.
  Sastry, ``Kalman filtering with intermittent observations,'' \emph{IEEE
  transactions on Automatic Control}, vol.~49, no.~9, pp. 1453--1464, 2004.

\bibitem{hoffmann}
G.~M. Hoffmann, S.~L. Waslander, and C.~J. Tomlin, ``Mutual information methods
  with particle filters for mobile sensor network control,'' in
  \emph{Proceedings of the 45th IEEE Conference on Decision and Control}.\hskip
  1em plus 0.5em minus 0.4em\relax IEEE, 2006, pp. 1019--1024.

\bibitem{mnih2013}
V.~Mnih, K.~Kavukcuoglu, D.~Silver, A.~Graves, I.~Antonoglou, D.~Wierstra, and
  M.~Riedmiller, ``Playing atari with deep reinforcement learning,''
  \emph{arXiv preprint arXiv:1312.5602}, 2013.

\bibitem{schulman2015trust}
J.~Schulman, S.~Levine, P.~Abbeel, M.~Jordan, and P.~Moritz, ``Trust region
  policy optimization,'' in \emph{International conference on machine
  learning}, 2015, pp. 1889--1897.

\bibitem{machado18}
M.~C. Machado, M.~G. Bellemare, E.~Talvitie, J.~Veness, M.~Hausknecht, and
  M.~Bowling, ``Revisiting the arcade learning environment: Evaluation
  protocols and open problems for general agents,'' \emph{Journal of Artificial
  Intelligence Research}, vol.~61, pp. 523--562, 2018.

\bibitem{charrow14}
B.~Charrow, V.~Kumar, and N.~Michael, ``Approximate representations for
  multi-robot control policies that maximize mutual information,''
  \emph{Autonomous Robots}, vol.~37, no.~4, pp. 383--400, 2014.

\bibitem{bu2008comprehensive}
L.~Bu, R.~Babu, B.~De~Schutter \emph{et~al.}, ``A comprehensive survey of
  multiagent reinforcement learning,'' \emph{IEEE Transactions on Systems, Man,
  and Cybernetics, Part C (Applications and Reviews)}, vol.~38, no.~2, pp.
  156--172, 2008.

\bibitem{omidshafiei2017deep}
S.~Omidshafiei, J.~Pazis, C.~Amato, J.~P. How, and J.~Vian, ``Deep
  decentralized multi-task multi-agent reinforcement learning under partial
  observability,'' in \emph{Proceedings of the 34th International Conference on
  Machine Learning-Volume 70}.\hskip 1em plus 0.5em minus 0.4em\relax JMLR.
  org, 2017, pp. 2681--2690.

\bibitem{lanctot2017unified}
M.~Lanctot, V.~Zambaldi, A.~Gruslys, A.~Lazaridou, K.~Tuyls, J.~P{\'e}rolat,
  D.~Silver, and T.~Graepel, ``A unified game-theoretic approach to multiagent
  reinforcement learning,'' in \emph{Advances in Neural Information Processing
  Systems}, 2017, pp. 4190--4203.

\bibitem{goodfellow2014generative}
I.~Goodfellow, J.~Pouget-Abadie, M.~Mirza, B.~Xu, D.~Warde-Farley, S.~Ozair,
  A.~Courville, and Y.~Bengio, ``Generative adversarial nets,'' in
  \emph{Advances in neural information processing systems}, 2014, pp.
  2672--2680.

\bibitem{baker2019emergent}
B.~Baker, I.~Kanitscheider, T.~Markov, Y.~Wu, G.~Powell, B.~McGrew, and
  I.~Mordatch, ``Emergent tool use from multi-agent autocurricula,''
  \emph{arXiv preprint arXiv:1909.07528}, 2019.

\end{thebibliography}
\vspace{16pt}

{\small
\begin{wrapfigure}{l}{1in}
\vspace{-16pt}
\begin{center}
    \includegraphics[width=1in]{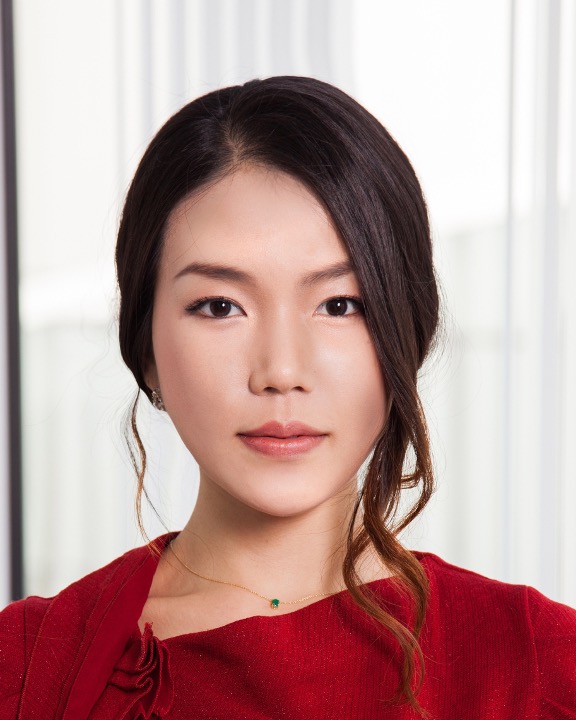}
\end{center}
\vspace{-16pt}
\end{wrapfigure}
\textbf{Heejin Jeong} (S'14) received her Ph.D. in Electrical ans Systems Engineering in 2020 and her M.S.E. degree in Robotics in 2016 at the University of Pennsylvania. She received her B.S. degree in Physics at Korea Advanced Institute of Science and Technology (KAIST) in 2014.

During her Ph.D., she worked at Amazon Robotics LLC as a research scientist intern in 2017 and at Waymo LLC as a software engineer intern in 2019 where she will continue to work as a software engineer after receiving her Ph.D.
Her research interests include reinforcement learning and its application to various automation such as robotics control. 

\vspace{16pt}
\begin{wrapfigure}{l}{1in}
\vspace{-16pt}
\begin{center}
    \includegraphics[width=1in]{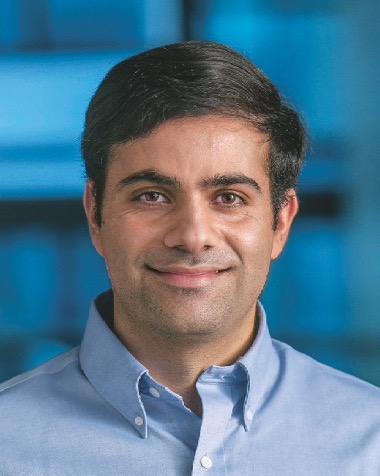}
\end{center}
\vspace{-16pt}
\end{wrapfigure}
\textbf{Hamed Hassani} (S'09-M'13) received a Ph.D. degree in Computer and Communication Sciences from EPFL, Lausanne.

He is currently an assistant professor of Electrical and Systems Engineering department at the University of Pennsylvania. Prior to that, he was a research fellow at Simons Institute for the Theory of Computing (UC Berkeley) affiliated with the program of Foundations of Machine Learning, and a post-doctoral researcher in the Institute of Machine Learning at ETH Zurich.

Dr. Hassani is the recipient of the 2014 IEEE Information Theory Society Thomas M. Cover Dissertation Award, 2015 IEEE Jack Keil Wolf ISIT Student Paper Award, 2017 Simons-Berkeley Fellowship, 2018 NSF-CRII Research Initiative Award, 2020 Air Force Office of Scientific Research (AFOSR) Young Investigator Award, and 2020 National Science Foundation (NSF) CAREER Award.

\vspace{16pt}
\begin{wrapfigure}{l}{1in}
\vspace{-16pt}
\begin{center}
    \includegraphics[width=1in]{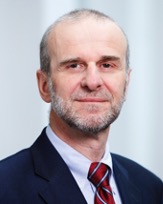}
\end{center}
\vspace{-16pt}
\end{wrapfigure}
\textbf{Manfred Morari} obtained the diploma from ETH Zurich and the Ph.D. from the University of Minnesota, both in chemical engineering.

He is currently a Distinguished Faculty Fellow at the University of Pennsylvania. Prior to this appointment, he was professor and head of Department of Information Technology and Electrical Engineering at ETH Zurich, the McCollum-Corcoran Professor of Chemical Engineering and Executive Officer for Control and Dynamical Systems at Caltech, and Professor at the University of Wisconsin.

Dr. Morari was president of the European Control Association. He supervised more than 80 Ph.D. students. He has received numerous awards, including Eckman, Ragazzini and Bellman Awards from AACC; Colburn, Professional Progress and CAST Division Awards from AIChE; Control Systems Award and Bode Lecture Prize from IEEE; Nyquist Lectureship and Oldenburger Medal from ASME; IFAC High Impact Paper Award. He is a Life Fellow of IEEE, and a Fellow of AIChE and IFAC, member of the U.S. National Academy of Engineering and Fellow of the UK Royal Academy of Engineering. Morari served on the technical advisory boards of several major corporations.

\vfill\null

\begin{wrapfigure}{l}{1in}
\vspace{-6pt}
\begin{center}
    \includegraphics[width=1in]{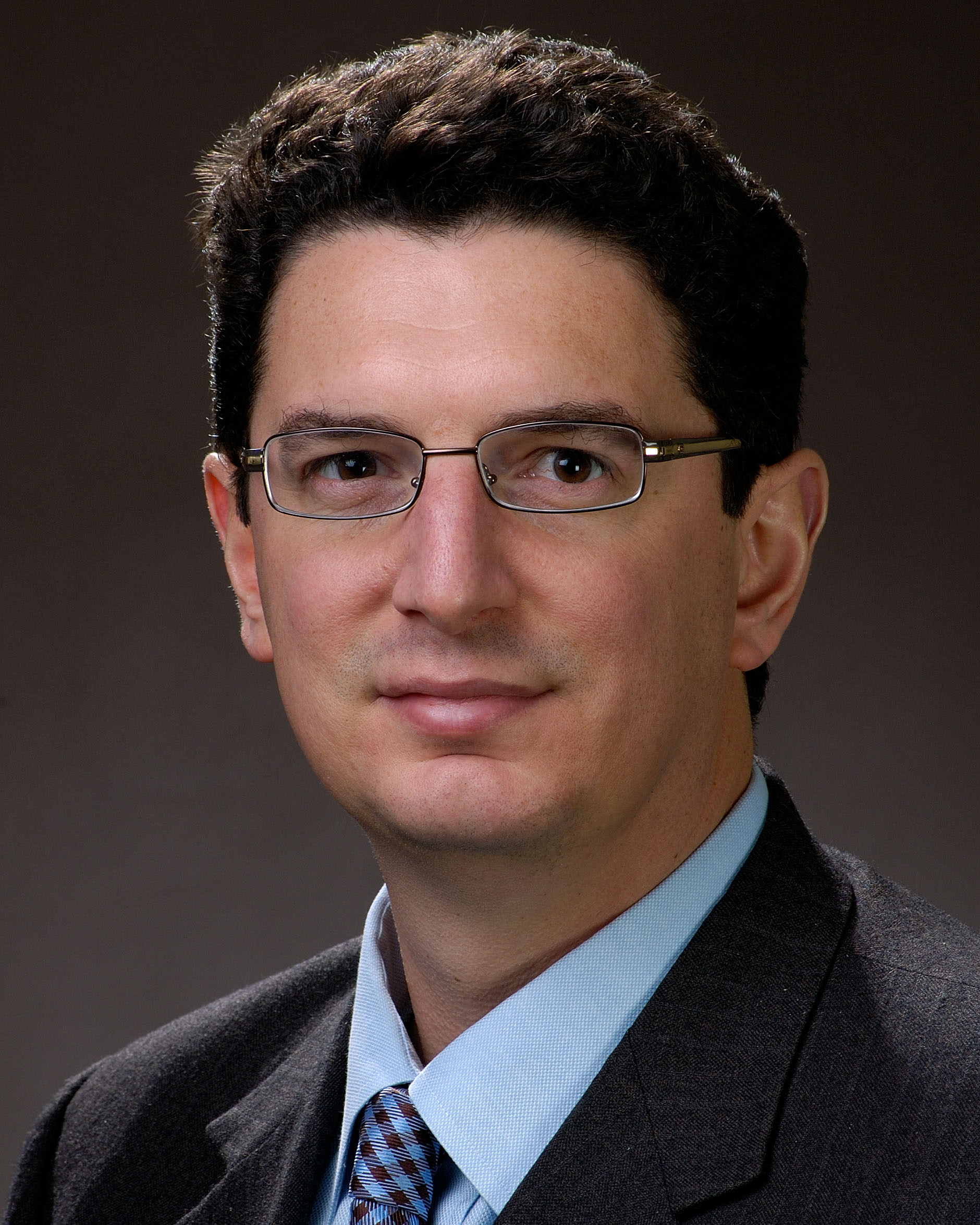}
\end{center}
\vspace{-16pt}
\end{wrapfigure}
\textbf{George J. Pappas} (S'90-M'91-SM'04-F'09) received the Ph.D. degree in electrical engineering and computer sciences from the University of California,
Berkeley, CA, USA, in 1998.

He is currently the Joseph Mis the UPS Foundation Professor and Chair of the Department of Electrical and Systems Engineering at the University of Pennsylvania. He also holds a secondary appointment in the Departments of Computer and Information Sciences, and Mechanical Engineering and Applied Mechanics. He is member of the GRASP Lab and the PRECISE Center. He has previously served as the Deputy Dean for Research in the School of Engineering and Applied Science. His research focuses on control theory and in particular, hybrid systems, embedded systems, hierarchical and distributed control systems, with applications to unmanned aerial vehicles, distributed robotics, green buildings, and biomolecular networks. 

Dr. Pappas is a Fellow of IEEE, an IFAC Fellow, and has received various awards such as the Antonio Ruberti Young Researcher Prize, the George S. Axelby Award, the O. Hugo Schuck Best Paper Award, the National Science Foundation PECASE, and the George H. Heilmeier Faculty Excellence Award.

\vspace{16pt}
\begin{wrapfigure}{l}{1in}
\vspace{-16pt}
\begin{center}
    \includegraphics[width=1in]{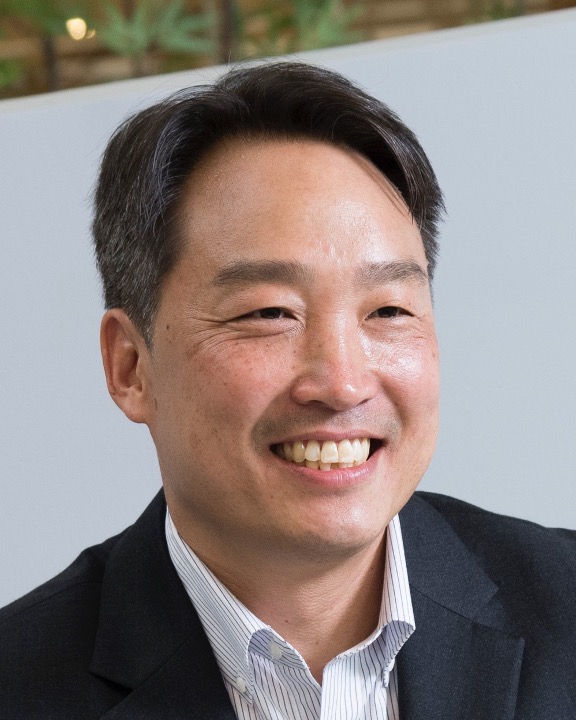}
\end{center}
\vspace{-16pt}
\end{wrapfigure}
\textbf{Daniel D. Lee} (M’12-SM’13-F’14) received his B.A. summa cum laude in Physics from Harvard University and his Ph.D. in Condensed Matter Physics from the Massachusetts Institute of Technology.

He is currently the Tisch University Professor in Electrical and Computer Engineering at Cornell Tech and Executive Vice President for Samsung Research. After completing his studies, he was a researcher at AT\&T and Lucent Bell Laboratories in the Theoretical Physics and Biological Computation departments. His research focuses on understanding general computational principles in biological systems, and on applying that knowledge to build intelligent robotic systems that can learn from experience.

Dr. Lee is a Fellow of the IEEE and AAAI and has received the National Science Foundation CAREER award and the Lindback award for distinguished teaching. He was also a fellow of the Hebrew University Institute of Advanced Studies in Jerusalem, an affiliate of the Korea Advanced Institute of Science and Technology, and organized the US-Japan National Academy of Engineering Frontiers of Engineering symposium and Neural Information Processing Systems (NeurIPS) conference.
}
\end{document}